\pdfoutput=1

\documentclass[11pt]{article}

\usepackage{acl}
\usepackage{times}
\usepackage{latexsym}
\usepackage{enumitem}

\usepackage[disable]{todonotes}

\usepackage[T1]{fontenc}
\usepackage[utf8]{inputenc}
\usepackage{microtype}
\usepackage{graphicx}
\usepackage{arydshln}
\usepackage[normalem]{ulem}

\title{Embarrassingly Easy Document-Level MT Metrics: \\How to Convert Any Pretrained Metric Into a Document-Level Metric}

\author{Giorgos Vernikos\thanks{{ }{ }Work conducted during an internship at Amazon.} \\
  EPFL + HEIG-VD \\
  georgios.vernikos@epfl.ch
  \\\And
  Brian Thompson \\
  AWS AI Labs \\
  brianjt@amazon.com
  \\\And
  Prashant Mathur \\
  AWS AI Labs \\
  pramathu@amazon.com 
 \\\And
 Marcello Federico \\
 AWS AI Labs \\
 marcfede@amazon.com \\
}

\begin{document}
\maketitle
\begin{abstract}
We hypothesize that existing sentence-level machine translation (MT) metrics become less effective  
when the human reference contains ambiguities. To verify this hypothesis, 
we present a very simple method for extending pretrained metrics to incorporate context at the document level. 
We apply our method to three popular metrics, BERTScore, Prism, and COMET, and to the reference free metric COMET-QE.
We evaluate the extended metrics on the WMT 2021 metrics shared task using the provided MQM annotations. 
Our results show that the extended metrics outperform their sentence-level counterparts in about 
85\% of the tested conditions, when excluding results on low-quality human references.
Additionally, we show that our document-level extension of COMET-QE dramatically improves its accuracy on discourse phenomena tasks, outperforming a dedicated baseline by up to 6.1\%.
Our experimental results support our initial hypothesis and show that a simple extension of the metrics permits them to 
take advantage of context to resolve ambiguities in the reference. 
\end{abstract}

\section{Introduction}
Automatic evaluation is crucial to the machine translation (MT) community
for tracking progress, evaluating new ideas and making modeling choices. 
While human evaluation is the gold standard for MT evaluation, it is very expensive, and
thus most research groups must rely on automatic metrics.
Current State-of-the-art (SOTA) metrics are \textit{pretrained} \cite{kocmi-etal-2021-ship, freitag-etal-2021-results}, leveraging existing language models or sequence-to-sequence models to judge how well a hypothesis (i.e. MT system output) conveys the same meaning as a human reference translation. 

Sentences are often ambiguous, and 
many recent works have demonstrated that incorporating inter-sentential (i.e. document-level) context is beneficial in both machine translation~\cite{lopes-etal-2020-document, fernandes-etal-2021-measuring} and human evaluation of MT~\cite{laubli-etal-2018-machine, toral-2020-reassessing, freitag-etal-2021-experts}.

A human reference translation is (at least ideally) created 
taking the entire source document into account.
However, just as source sentences are often ambiguous, 
we hypothesize that human reference sentences also contain ambiguities. 
Thus, when a system output deviates from the human reference, 
we may need to look at additional context to determine if those deviations are acceptable, in the context of the full document translation. \todo{put tables on correct pages!}

In this study, we present a simple procedure for extending pretrained MT metrics to the document level.
Prior work has used pretrained models (e.g. BERT) to embed a single human reference sentence and hypothesis (e.g. an MT output) sentence.
We instead argue that a \textit{better} representation of the reference or hypothesis sentence can be obtained by providing several sentences of context to the pretrained model, 
allowing the pretrained model to \textit{use surrounding context when embedding each sentence of interest}. 
Once the embeddings of the reference or hypothesis sentence have been computed (taking into account surrounding sentence context), the metric is computed in the same manner as the sentence-level metric.\footnote{In the case of Prism~\cite{thompson-post-2020-automatic}, 
we modify this logic slightly to retain only the probabilities of the sentence of interest (see \autoref{method:prism}).}\textsuperscript{,}\footnote{In the case of COMET/COMET-QE~\cite{rei-etal-2020-comet}, which incorporates the source sentence, we provide additional source context in the same manner (see \autoref{method:comet}).}

\vspace{2mm}

We apply this method to extend four popular pretrained metrics to the document level:\footnote{We will release our code upon acceptance.}
\begin{itemize}[noitemsep,topsep=0pt]
    \item BERTScore~\cite{Zhang*2020BERTScore:}, a text generation metric that uses the alignments from token embeddings of a pretrained BERT model to score an MT output given the reference
    \item Prism~\cite{thompson-post-2020-automatic}, a text generation metric which utilizes a sequence-to-sequence paraphrase model to estimate the probability that an MT output is a paraphrase of the reference translation
    \item COMET~\cite{rei-etal-2020-comet}, an MT metric which fine-tunes a multilingual language model, namely XLM-R~\cite{conneau-etal-2020-unsupervised}, to predict human translation quality judgments. 
    \item COMET-QE~\cite{rei-etal-2020-comet}, the reference-free (i.e. ``quality estimation as a metric'') version of COMET
\end{itemize}

To test the effectiveness of our document-level metrics, we measure system-level correlation with human judgments.
We select the so-called "platinum" Multidimensional Quality  Metrics (MQM) judgments collected for the WMT 2021 metrics task \cite{freitag-etal-2021-results}.
We believe MQM judgments are the best available to test document-level MT metrics as
these judgments are
made by expert translators 
that have access to,
and are strongly advised to consider,
source-side document-level context when judging each target sentence. 
We perform evaluation on all the WMT 2021 language pairs (En$\rightarrow$De, Zh$\rightarrow$En, En$\rightarrow$Ru) and domains (TED talks and news) for which MQM judgments are available.

We find that our document-level extensions of these four metrics outperform their sentence-level counterparts in 75\% of cases considered.
Excluding Zh$\rightarrow$En news, where the human reference is of low quality (see \autoref{data:mqm}), we see improvements in 85\% of cases. 
This provides strong evidence that document-level context is useful in the automatic MT evaluation.

We also conduct analysis to better understand the performance improvement that we observe.
We show that using reference context is better than using context from the MT output, likely because the MT output contains more errors than the reference. 
We also demonstrate that
our document-level extension of COMET-QE significantly improves over its sentence-level counterpart on
targeted tasks evaluating discourse phenomena,
namely pronoun resolution 
and word sense disambiguation.\footnote{The use of a reference would make these tasks trivial, so 
we limit our analysis to the reference-free COMET-QE.} 
This finding provides further evidence that our document-level metrics 
are using context to disambiguate ambiguities in the reference sentence. 

In summary, our contributions are:
\begin{enumerate}[noitemsep,topsep=0pt]
    \item We present a simple but effective method to extend pretrained sentence-level metris to the document level, and apply it to four popular metrics. 
    \item  We show that the proposed document-level metrics tend to have better correlation with human judgments than their sentence-level counterparts.
    \item  We improve over both COMET and COMET-QE, which appear to be the previous SOTA automatic metric and reference-free metric, respectively \cite{freitag-etal-2021-results, kocmi-etal-2021-ship}.
    \item We conduct analysis to show that the improvements observed using our approach can be attributed to better context utilization, and also show that using reference context is better than using context from the hypothesis.

\end{enumerate}

\section{Related Work}
Our work has parallels in human MT evaluation, where 
document-level judgments are required to 
distinguish human translation quality from MT system quality
\cite{laubli-etal-2018-machine, toral-2020-reassessing}. 
\citet{castilho-etal-2020-context} showed that many source sentences are ambiguous, but that ambiguities are often resolved using only a few additional sentences of context. This suggests that we do not need to integrate very many additional sentences of context into a document-level metric in order to see an improvement in quality.

Pretrained metrics are metrics 
which leverage large existing pretrained language models or sequence-to-sequence models, and include 
YiSi \cite{lo-2019-yisi},
COMET~\cite{rei-etal-2020-comet}, BERTscore~\cite{Zhang*2020BERTScore:}, Prism~\cite{thompson-post-2020-automatic}, BLEURT~\cite{sellam-etal-2020-bleurt}, and others.
Pretrained metrics have been shown to consistently outperform surface-level metrics such as BLEU \cite{papineni-etal-2002-bleu}, TER \cite{snover-etal-2006-study}, and chrF \cite{popovic-2015-chrf} -- see \citet{mathur-etal-2020-results, kocmi-etal-2021-ship, freitag-etal-2021-results}. 

\begin{figure*}[ht]
    \centering
    \includegraphics[width=\textwidth]{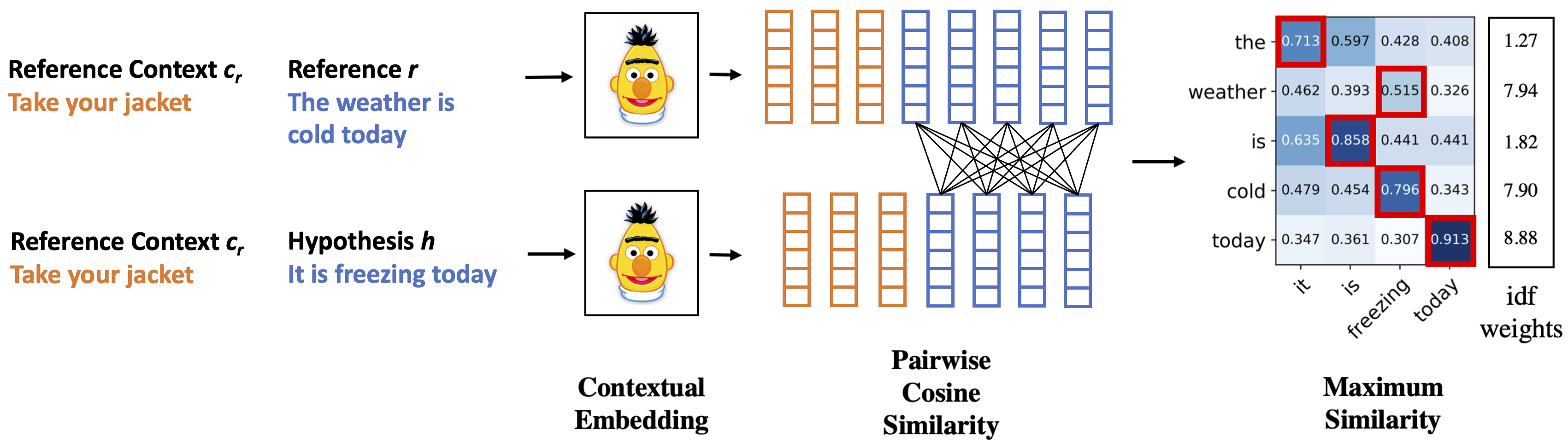}
    \caption{To extend BERTScore to the document level, we add reference context to both the reference sentence and the hypothesis sentence. 
    This context is used to improve the embeddings of the reference and hypothesis sentences, but is not used when performing alignment and scoring, which follows standard sentence-level BERTScore. 
    The same methodology is applied to Prism and COMET/COMET-QE (not shown). Image adapted from \citet{Zhang*2020BERTScore:}.}
    \label{fig:method}
\end{figure*}

Prior to the rise of pretrained metrics, 
several works targeted discourse-level phenomena in MT metrics such as pronominal anaphora \cite{hardmeier-federico-2010-modelling,jwalapuram-etal-2019-evaluating}
and lexical cohesion \cite{wong-kit-2012-extending, gong-etal-2015-document}.
For a detailed overview of evaluation of discourse-level phenomena, we direct the reader to \cite{Sameen2021Survey}. 
Recently, \citet {jiang-etal-2022-blonde} proposed BlonDe, a document-level metric that focuses on discourse phenomena in order to score a translated document. However, 
the authors did not compare to modern, pretrained metrics and we find that BlonDe substantially under-performs such metrics, despite taking advantage of document-level context (see discussion in Sec \ref{sec:experiments}).

\section{Method}

At a high level, we follow a very simple procedure for extending pretrained MT metrics to the document level: 
As in standard sentence-level metrics, we produce a score for a single hypothesis sentence compared to a single human reference translation sentence.
However, we use additional context (prior sentences) from the reference translation when computing the contextual embeddings for both the hypothesis and reference sentence.
Once the hypothesis and reference sentence have been embedded, we discard the extra context sentences before computing metric scores following the same process as in the corresponding sentence-level metric. 
Additional details are provided for each metric below. 

For the following discussion, let $s$ refer to the source sentence,
$h$ refer to the hypothesis (i.e. MT system output) sentence,
$r$ refer to the human reference translation sentence, and
let $ c_s $, $ c_h $ and $ c_r $
refer to the source, hypothesis, and reference context, respectively. 

\subsection{Document-level BERTScore}\label{method:bertscore}
BERTScore~\cite{Zhang*2020BERTScore:} is an unsupervised text generation metric that leverages the power of a pretrained large Language Model (LM) to score generated text. BERTScore encode tokens of both the reference and the hypothesis with a pretrained LM and computes soft alignments based on token similarities. The alignment matrix is then used to calculate the precision, recall and F1 scores of the hypothesis compared to the reference.

To extend BERTScore to the document level, we concatenate the reference context $\langle c_r \rangle$ while encoding the hypothesis or the reference with the LM. 
However, we align only the tokens of the reference/hypothesis sentence being scored 
(see \autoref{fig:method} for an illustration).

For BERTScore we use the default LM option for each language pair, which is the multilingual BERT-base~\cite{devlin-etal-2019-bert} for all En$\rightarrow$* pairs and RoBERTa-large~\cite{roberta} for *$\rightarrow$En pairs. 
BERT and RoBERTa are naively document-level; specifically, the LMs are trained on up to 512 tokens at a time, which is significantly longer than the average sentence length.
Thus no changes to the underlying model were required to extend BERTscore to the document level. 

\subsection{Document-level Prism}\label{method:prism}
Prism~\cite{thompson-post-2020-automatic} is an unsupervised text generation metric that uses a sequence-to-sequence paraphraser to evaluate how well a hypothesis paraphrases a human reference translation. 
Specifically, to score a translation the reference is fed to the encoder and the hypothesis is force-decoded in the decoder via teacher forcing. 
The token-level probabilities of the reference are aggregated to produce a score and the process is repeated with the hypothesis in the encoder side and the reference in the decoder. The final score is the average of the two scores.

In order to generalize Prism for document-level evaluation we concatenate the reference context $ c_r $ to both the encoder and decoder $\langle c_r;r, c_r;h \rangle$. The context fed in the decoder serves as a prompt and we only aggregate token-level probabilities for the sentence being evaluated.

The authors of Prism release a sentence-level multilingual MT model for use as a zero-shot paraphrase model.
One option for extending Prism to the document level is to train a document-level, multilingual MT model.
While document-level 
data collection methods and datasets do exist~\cite{guo-etal-2019-hierarchical, thompson-koehn-2020-exploiting, cettolo-etal-2012-wit3, lison-etal-2018-opensubtitles2018}, document-level data is not currently available in nearly as many language pairs as sentence-level data.
To extend Prism to the document level, we instead use  mBART-50~\cite{Tang2020MultilingualTW},
a multilingual encoder-decoder language model. 
mBART-50 is trained on document fragments of up to 512  tokens, in 50 languages,
resulting in a multilingual document-level paraphraser.\footnote{While  mBART can be fine-tuned on (sentence-level) translation, we do not do so because we require the model to work at the document level. Additionally, although mBART-50 is multilingual, it is not a translation model so we cannot use it for the reference-free version of Prism.}

\subsection{Document-level COMET}\label{method:comet}
COMET~\cite{rei-etal-2020-comet} is a supervised metric that is trained on human judgments. COMET encodes the source, hypothesis and reference via a multilingual pretrained LM and the representation of each sentence is the average of its output token embeddings. 
The encoded representations are further combined via subtraction and multiplication and fed to a regressor that predicts a score for each translated sentence. 
We use the latest COMET-MQM\_2021 
model~\cite{rei-etal-2021-references} that is built on top of XLM-RoBERTa-large~\cite{conneau-etal-2020-unsupervised}. The models are pretrained on Direct Assessments from WMT 2015 to WMT 2020 and fine-tuned on MQM z-scores from \citet{freitag-etal-2021-experts}. 

To extend COMET to the document level, we integrate  source $ c_s $ and reference context $ c_r $ by concatenating them with the source and hypothesis/reference in the encoder. 
We obtain sentence representations by averaging the output embeddings of the tokens of the current sentence only before passing them to the regressor.

As with BERTscore, the model underlying COMET is inherently document-level.
However, the underlying LM is fine-tuned for a few epochs on human judgments from previous WMT campaigns that consist of a single (source, reference, and hypothesis) sentence and the corresponding score. 
As the amount of fine-tuning is quite limited,
we hypothesize that the model has still retained its ability to handle text beyond sentence level, and this assumption appears to be confirmed by experimental results (see \autoref{results:correlation}). 

\subsection{Document-level COMET-QE}
COMET-QE~\cite{rei-etal-2021-references} is the reference-free version of COMET.
We use the latest COMET-MQM-QE\_2021, trained similarly to the COMET-MQM\_2021 discussed above.
Although COMET-QE does not does not have access to the reference it has been shown to perform reasonably well compared to strong reference-based metrics~\cite{kocmi-etal-2021-ship}.

Similar to reference-based COMET, to extend COMET-QE to the document level, 
for each source $ s $ and hypothesis $ h $, we concatenate the previous source and hypothesis sentences as context $\langle c_s;s, c_h;h \rangle$ and score the hypothesis $ h $ in question. 

The pretrained model for COMET-QE is the same as the one used in COMET, therefore no further modifications are required to extend COMET to the document level.

\begin{table*}[ht]
\centering
\resizebox{\textwidth}{!}{
\begin{tabular}{l|c|ccc|ccc}
\hline
\textbf{Model} & \textbf{Input} & \multicolumn{3}{c|}{\textbf{TED talks}} & \multicolumn{3}{c}{\textbf{News}} \\ \hline
 & & En$\rightarrow$De & En$\rightarrow$Ru & Zh$\rightarrow$En & En$\rightarrow$De & En$\rightarrow$Ru & Zh$\rightarrow$En \\ \hline
  BlonDe & $\langle c_h, h, c_r, r \rangle$ & - & - & -0.232 & - &  - & 0.212 \\ 
 Prism (m39v1) & $\langle h, r \rangle$ & 0.656 & 0.867 & 0.272 & 0.841 & 0.799 & 0.558 \\ \hline 
Prism (mBART-50) & $\langle h, r \rangle$ & 0.486 & 0.845 & 0.240 & 0.661  & 0.710 & 0.363 \\
Doc-Prism (mBART-50) & $\langle c_r;h, c_r;r \rangle$ & \textbf{0.692}  & \textbf{0.852} & \textbf{0.372} & \textbf{0.825}\textsuperscript{*} & \textbf{0.777} & \textbf{0.374} \\ \hline
BERTScore & $\langle h, r \rangle$ & 0.506 & 0.831 & 0.293 & 0.930  & \textbf{0.629} & \textbf{0.575}\textsuperscript{*} \\ 
Doc-BERTScore & $\langle c_r;h, c_r;r\rangle$ & \textbf{0.613}\textsuperscript{*}  & \textbf{0.836} & \textbf{0.344}\textsuperscript{*} & \textbf{0.948}\textsuperscript{*}  & 0.622 & 0.535 \\ \hline
COMET & $\langle s, h, r\rangle$ & \textbf{0.818}  & 0.841& 0.266  & 0.772  & 0.659 & \textbf{0.628} \\ 
Doc-COMET\ & $\langle c_s;s, c_r;h, c_r;r \rangle$ & 0.816  & \textbf{0.849}& \textbf{0.297} & \textbf{0.802\textsuperscript{*}}  & \textbf{0.676}& 0.513 \\ \hline
COMET-QE & $\langle s, h \rangle$ & 0.694  & 0.818 &\textbf{ -0.209} & 0.711  & 0.688&  \textbf{0.529} \\
Doc-COMET-QE & $\langle c_s;s, c_h;h \rangle$ & \textbf{0.724}  & \textbf{0.830}& -0.255 & \textbf{0.733} & \textbf{0.733}\textsuperscript{*}  & 0.462 \\ \hline
\end{tabular}
}
\caption{System-level Pearson correlation with WMT 2021 MQM annotations for Prism, BERTScore, COMET and COMET-QE and their generalization for document-level evaluation (Doc-*, this work). 
Within each document/sentence-level pair, \textbf{bold} denotes the best correlation and ``*'' denotes a statistically significant ~($p < 0.05$) difference. 
Excluding Zh$\rightarrow$En news data, which has a very low-quality human reference (see \autoref{data:mqm}), our document-level metrics outperform their sentence-level counterparts in 17 of 20 (85\%) of cases, and 6 of 6 (100\%) of statistically significantly different cases. 
}
\label{tab:results}
\end{table*}

\section{Experiments}
\label{sec:experiments}

Motivated by the finding of \citet{scherrer-etal-2019-analysing,kim-etal-2019-document,castilho-etal-2020-context} that two previous sentences are sufficient context to correctly resolve ambiguities in
the majority of sentences,
we use two previous reference sentences as context unless otherwise noted. 
Sentences are separated using the separator token of each model: [SEP] for RoBERTa and <\textbackslash s> for XLM-R and mBART-50.
We use reference context $ c_r $ as reference for the hypothesis, as opposed to 
hypothesis context $ c_h $.
This is done in order to avoid propagation of translation errors (see \autoref{sec:analysis:ref} for an ablation using hypothesis context instead of reference context).

\subsection{Human Judgment Experiments}\label{data:mqm}

We compare our document-level metrics judgments of MT outputs with those of the human-generated 
MQM annotations from the 2021 WMT metrics shared task~\cite{freitag-etal-2021-experts}. We select MQM over Direct Assessment annotations for two reasons: First, they are produced by professional translators (compared to crowd workers or translation researchers) and require explicit error annotations that should lead to higher quality annotations. Second, MQM annotators are specifically instructed to take context into account "\textit{...identify all errors within each segment in a document, paying particular attention to document context}" which is the best scenario for document-level evaluation. In 2021, in addition to the news domain, annotations were also provided for translations of TED talks in three language pairs: En$\rightarrow$De, Zh$\rightarrow$En and En$\rightarrow$Ru. 

One potential problem with the metrics dataset is the quality of the 
Zh$\rightarrow$En news human reference.
The WMT metrics shared task organizers acquired MQM scores for the human references, in addition to MT system outputs. The Zh$\rightarrow$En 
reference received an MQM score of just 4.27, only slightly better than the best MT system at 4.42 \cite{freitag-etal-2021-results}. For reference, a score of 5 corresponds to one major error per sentence.
In contrast, for the same language pair, the TED reference has an MQM score of 0.42 vs the best MT system at 1.65.

\subsection{Discourse Phenomena Experiments}

In order to 
confirm that any gains we see from document-level metrics are in fact due to their ability 
to correctly handle ambiguities in the reference which can be resolved using document-level context,
we also perform targeted evaluation of discourse phenomena using contrastive sets. 
These testsets are common in the evaluation of document-level MT systems where a context-aware model should ideally assign the highest probability to the correct translation;
all translations are plausible and only the use of context can reveal the correct translations. 
For our case, since we are evaluating MT metrics, we consider each sentence as a translation generated by a different model and calculate how often our metric ranks the correct translation the highest. 
Since the use of a reference would make this task trivial for reference-based metrics, we only evaluate on COMET-QE. 
We use ContraPro~\cite{muller-etal-2018-large}, 
a selection of sentences from OpenSubtitles2018~\cite{lison-etal-2018-opensubtitles2018} that  contain the English anaphoric pronoun \textit{it} in the source side. 
Starting from the correct translation in German, contrastive translations are automatically created by translating into the German pronouns \textit{er}, \textit{sie} and \textit{es}. 
In order to identify the correct translation the model must look into previous context. 
We also evaluate on a similar dataset for En$\rightarrow$Fr created by \citet{lopes-etal-2020-document} for the translation of \textit{it} and \textit{they} into \textit{il}, \textit{elle}, \textit{ils}, \textit{elles} in French.
Finally, we evaluate on DiscEvalMT~\cite{bawden-etal-2018-evaluating}, a contrastive test which consists of 200 examples of anaphoric pronoun translation for En$\rightarrow$Fr and  200  examples  of  word sense disambiguation (WSD).

\begin{table*}[ht]
    \centering
    \begin{tabular}{l|ccc|ccc|c|c} 
        \hline
        \textbf{Model} & \multicolumn{3}{c|}{En$\rightarrow$De} & \multicolumn{5}{c}{En$\rightarrow$Fr} \\ \hline
        & Intra & Inter & Total & Intra & Inter & Total &  Anaphora & WSD \\ \hline
        \citet{lopes-etal-2020-document} & - & - & 70.8 & - & - & 83.2 & 82.5 & 55.0 \\ \hline
        COMET-QE & 78.2 & 40.9 & 48.4 & 76.3 & 76.6 & 76.5 & 50.0 & 50.0 \\
        Doc-COMET-QE (this work) & \textbf{80.5} & \textbf{72.6} & \textbf{74.2} & \textbf{88.7} & \textbf{88.0} & \textbf{88.3} & \textbf{83.5} & \textbf{68.0}\\ \hline
    \end{tabular}
    \caption{Accuracy (percentage correct) for targeted evaluation of contextual phenomena. Our document-level version of COMET-QE substantially outperforms the sentence-level COMET-QE, and also outperforms the best methods proposed by \citet{lopes-etal-2020-document}, demonstrating that it is successfully incorporating contextual information.}
    \label{tab:targeted_eval}
\end{table*}

\subsection{Baseline Methods}

For correlation with human MT quality judgments, in addition to the sentence-level version of each metric we extend, we also compare to BlonDe~\cite{jiang-etal-2022-blonde}, an overlap-based document-level metric that focuses on discourse phenomena,\footnote{We report BlonDe results in English only, 
as BlonDe uses a discourse marker script from \citet{sileo-etal-2019-mining} which was trained only in English. BlonDe could likely be extended to other languages but we did not attempt to do so.}
and Prism using the m39v1 model released by the authors of Prism. 

For discourse phenomena, we compare our document-level COMET-QE model to the sentence-level COMET-QE as well as the best reported results of \citet{lopes-etal-2020-document}.

\section{Results}

\subsection{Correlation with Human Judgments}\label{results:correlation}

We present the system-level Pearson correlation with the human annotations of the 2021 WMT metrics task for all metrics (sentence- and document-level) in \autoref{tab:results}. 
Statistically significance ~($p < 0.05$) is computed for each sentence- vs document-level metric pair following \citet{freitag-etal-2021-results} using the PERM-BOTH hypothesis test \cite{perm-both/tacl_a_00417}.
We also provide the results of BlonDe 
(only for *->En since this metric relies on entity taggers and discourse markers that are only trained in English) 
and Prism with the original model (m39v1) for reference. 

Overall, 
adding document-level context leads to improved correlation with human judgments for all metrics. 
Our document-level metrics outperform their sentence-level counterparts in 18 of 24 (75\%) of cases considered. 
Excluding Zh$\rightarrow$En news data, which has a very low-quality human reference (see \autoref{data:mqm}), our document-level metrics outperform their sentence-level counterparts in 17 of 20 (85\%) of cases. 
Looking at only pairs with statistically significant differences,
our document-level metrics outperform their sentence-level counterparts in
6 of 7 cases (86\%), and 6 of 6 (100\%) of cases excluding Zh$\rightarrow$En news.

We see that document-level metrics outperform sentence-level metrics in only 1 of 4 cases on Zh$\rightarrow$En news
This suggests that the document-level metrics are sensitive to errors in the reference context. This hypothesis is further supported by analysis in \autoref{sec:analysis:ref}.

For Prism we observe that the sentence-level results with the original m39v1 model are better than with mBART-50 on every language pair/domain which shows mBART-50 is suboptimal when used with Prism.
However, using document-level context we are able to consistently improve in every language pair/domain over the Prism (mBART-50) and close the gap to Prism (m39v1) sentence-level model. Doc-Prism (with a suboptimal model) even outperforms the stronger m39v1 model in two TED language pairs.

Although the COMET models are fine-tuned on single sentence MT judgments, 
experimental results suggest they are able to retain their ability to handle intersentential dependencies.
We considered retraining COMET excluding older direct assessment judgments which did not take document-level context into account; however this would have severely limited the amount of (already very limited!) training data. 

Finally, we observe that BlonDe performs significantly worse than the pretrained metrics as well as our document-level extensions, underperforming everything except document-level COMET-QE in TED Zh$\rightarrow$En. 

\begin{table*}[ht]
    \centering
    \begin{tabular}{l|c|c|c|c} 
        \hline
     & \textbf{Context} & Doc-Prism & Doc-BERTScore & Doc-COMET \\ \hline
    hypothesis & $\langle c_s;s, c_r;r, c_h; h \rangle$ &         0.595  &         0.624  &         0.630 \\
    reference & $\langle c_s;s, c_r;r, c_r; h \rangle$ & \textbf{0.649} & \textbf{0.650} & \textbf{0.659} \\ \hline
    \end{tabular}
    \caption{Average correlation with MQM human judgments of our document-level metrics using previous hypothesis sentences as context vs. previous reference sentence as context. COMET-QE is excluded because it does not depend on the reference. For all three methods, we see better correlation using the reference for hypothesis context. We hypothesize that this is because using previous hypothesis sentences allows for propagation of errors (i.e. an error in a previous sentence can impair the judgment of the current sentence).}
    \label{tab:context_quality}
\end{table*}

\subsection{Discourse Phenomena Improvements}

We provide the results of targeted evaluation on contrastive datasets for COMET-QE and Doc-COMET-QE in \autoref{tab:targeted_eval}. 
We also provide the scores of the best-performing document-MT model for each dataset from \citet{lopes-etal-2020-document} for reference. 
The reference-based metrics are not considered in this section as the use of a reference would make the task trivial.

We observe that the document-level COMET-QE substantially outperforms the sentence-level COMET-QE, and even models optimized for document-level translation. 
Surprisingly, we observe improvements in the evaluation of pronoun translation not only when the necessary information is located in a previous sentence (Inter) but even in the case where the antecedent can be found in the same sentence (Intra), suggesting additional context is helpful in these cases as well. 
Apart from pronoun translation, our approach also improves over both the sentence-level metric and the document-level MT at word sense disambiguation. 
These findings 
all support our hypothesis that our document-level metrics are resolving ambiguities in the reference sentence by using additional context.

\section{Ablations}\label{sec:analysis}
\subsection{Hypothesis vs Reference Context}\label{sec:analysis:ref}
For our document-level MT metrics described prior to this point, we use the reference context $ c_r $ (as opposed to the hypothesis context $ c_h $) as context for the hypothesis. 
Our reasoning behind this decision is that previous translations could contain errors that might bias the document-level metric into rewarding erroneous translations. 
To test this, we conduct an ablation experiment in which we concatenate the hypothesis context to the hypothesis while the context of the remaining inputs, i.e. the reference and the source sentence, remains unchanged. \autoref{tab:context_quality} shows the average correlation across all language pairs and domains using either the hypothesis context or the reference context. We do not provide these scores for COMET-QE as it not have access to the reference. 

\begin{figure*}[t]
    \centering
    \includegraphics[scale=0.4]{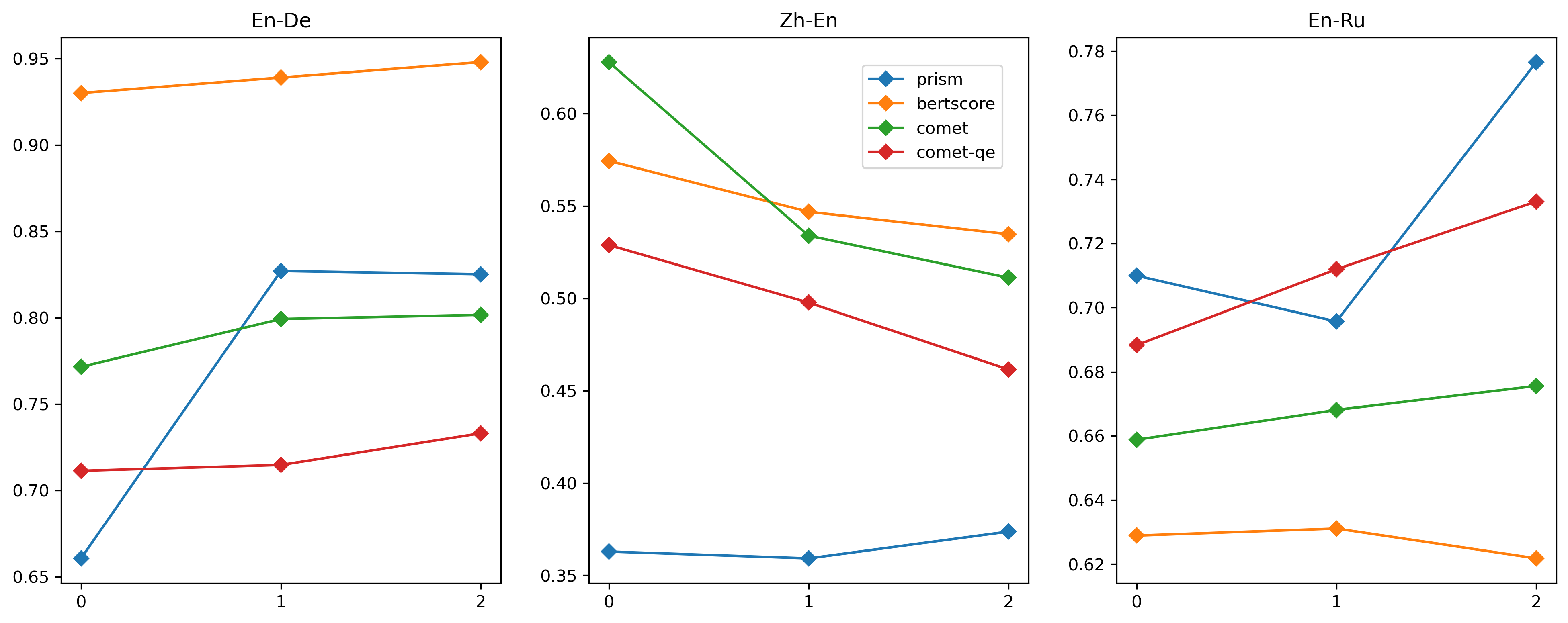}
    \includegraphics[scale=0.4]{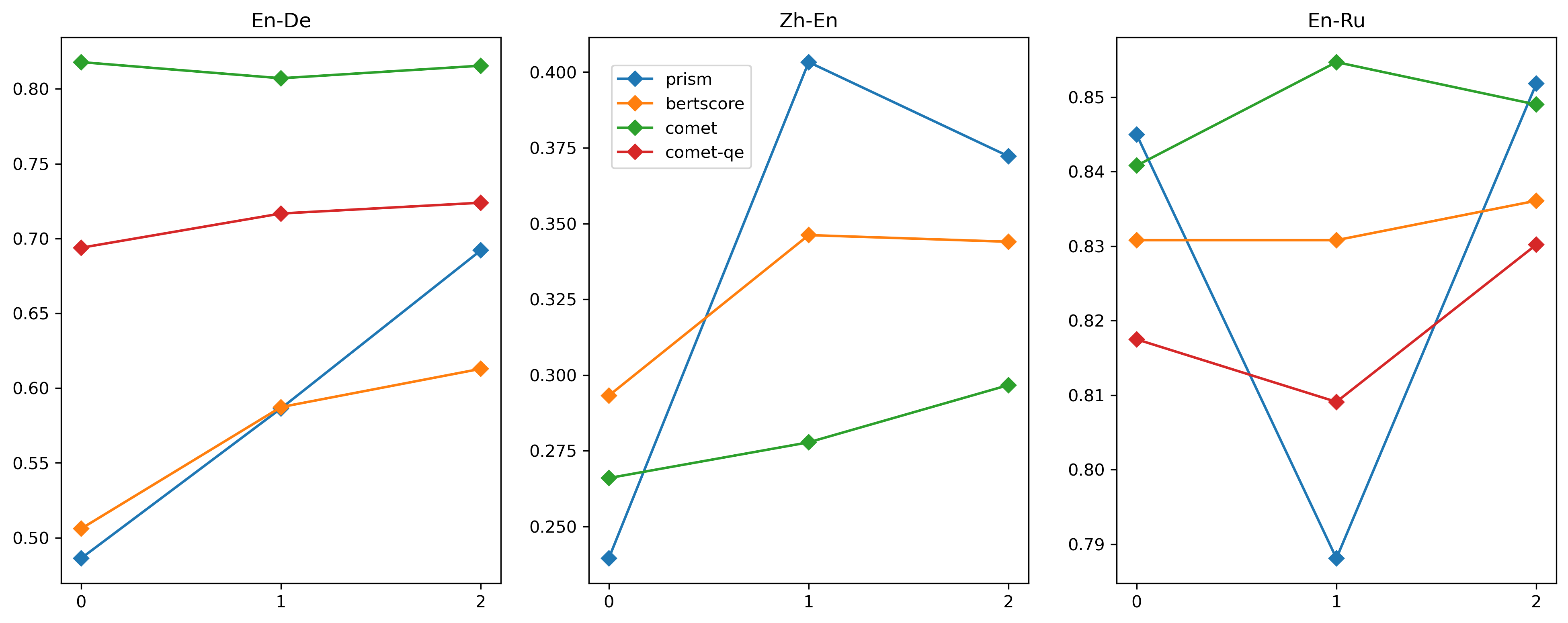}
    \caption{System-level Pearson correlation with human correlation vs. number of sentences of context for News (upper) and TED talks (lower). Although the results are noisy, in general we see correlation improve as the amount of context increases. The one exception is Zh$\rightarrow$En News, which we attribute to poor human references (see \autoref{data:mqm}).}
    \label{fig:context}
\end{figure*}

We observe that the use of the hypothesis context degrades performance for all metrics, 
which confirms the findings of \citet{fernandes-etal-2021-measuring} for document-level machine translation. 
We hypothesize that this is because 
the previous hypothesis sentences contain more errors than previous reference sentences, and thus using previous hypothesis sentences allows for more propagation of errors (i.e. an error in a previous sentence can impair the judgment of the current sentence). 

One disadvantage of using reference context for the hypothesis is that we cannot measure document-level fluency, that is, how well a document flows from one sentence to the next. Our analysis suggests that either document level fluency is of less concern than error propagation, and/or that MQM judgments are not adequately capturing document-level fluency.

\subsection{Amount of Context}
In our experiments so far we use the previous two sentences as context, motivated by the finding of \citet{scherrer-etal-2019-analysing,kim-etal-2019-document,castilho-etal-2020-context} that two previous sentences are sufficient context to 
resolve ambiguities in the majority of sentences. 
To assess the amount of context that is needed for document-level MT metrics we also perform evaluation with the previous sentence as context. 
\autoref{fig:context} shows the results for [0, 1, 2] previous sentences as context for news articles and TED talks. 
In the news domain we observe that for En$\rightarrow$De and En$\rightarrow$Ru), adding more context helps. On the other hand, for Zh$\rightarrow$En, adding context appears to be harmful. We believe this is likely explained by the relatively low-quality human references in Zh$\rightarrow$En (see \autoref{data:mqm}).
For TED talks,
although the results are somewhat noisy,
we also observe that context tends to improve correlation across all three language pairs.

\section{Conclusion}

We proposed a simple and effective approach to generalize pretrained MT metrics to the document level.
We apply our approach to BERTScore, Prism, COMET-QE, and COMET-QE, and we believe that it could easily be extended to other pretrained sentence-level metrics. 
To the best of our knowledge, our work is the first example of pretrained document-level MT metrics.

We demonstrate that the use of document-level context in pretrained metrics improves correlation with human judgments, and that the improvements are likely coming from better context utilization. 
We present results on MT evaluation but our approach may also be
benefits in other Natural Language Generation (NLG) tasks
where discourse phenomena 
are present (e.g paraphrasing, data to text generation, chatbots, etc). 
In conclusion, we argue that the 
MT community (and possibly the greater NLG community)
should adopt metrics such as those presented in this work which take document-level context into account, and which align better with the human evaluation process where annotators are encouraged to use document level context.
We also recommend that any future research in metrics should explore novel ways to incorporate context.

\clearpage

\bibliography{main.bbl}
\bibliographystyle{acl_natbib}

\end{document}